\documentclass{sig-alternate-2013}

\permission{Copyright is held by the International World Wide Web Conference Committee (IW3C2). IW3C2 reserves the right to provide a hyperlink to the author's site if the Material is used in electronic media.}
\conferenceinfo{WWW'14,}{April 7--11, 2014, Seoul, Korea.} 
\copyrightetc{ACM \the\acmcopyr}
\crdata{978-1-4503-2744-2/14/04. \\
http://dx.doi.org/10.1145/2566486.2567988
}
 
\usepackage{cmap}
\usepackage[OT1]{fontenc}

\usepackage{multirow}
\usepackage{booktabs}
\usepackage{microtype}
\usepackage{subfigure}
\usepackage{balance}
\usepackage[boxed, linesnumbered , ruled]{algorithm2e}

\usepackage[svgnames,rgb,usenames,dvipsnames,svgnames,table]{xcolor}
\usepackage[pdfpagemode={UseOutlines},
	bookmarks,
	bookmarksopen=false,
	pdfstartview={FitH},
	bookmarksnumbered=true,
	bookmarksnumbered,
	colorlinks,
	pdftex,
	pdfpagelabels=false,
	plainpages=false,
	linkcolor = {RoyalBlue},
	citecolor = {Maroon},
    urlcolor = {ForestGreen}]{hyperref}

\begin{document}

\title{Quizz: Targeted Crowdsourcing\\with a Billion (Potential) Users}

\numberofauthors{2}
 
\author{
%\alignauthor
%Panos Ipeirotis \qquad Evgeniy Gabrilovich\\
%\affaddr{Google, Mountain View, CA 94043}
%\email{\{ipeirotis|gabr\}@google.com}
%
\alignauthor
Panagiotis G.\ Ipeirotis\thanks{Work done while visiting Google.}\\
       \affaddr{Google \& New York University}\\
%       \affaddr{1600 Amphitheatre Parkway}\\
       %\affaddr{Mountain View, CA 94043}\\
       \email{panos@stern.nyu.edu}
\alignauthor
Evgeniy Gabrilovich\\
       \affaddr{Google}\\
%       \affaddr{1600 Amphitheatre Parkway}\\
       %\affaddr{Mountain View, CA 94043}\\
       \email{gabr@google.com}
}

\maketitle

\begin{abstract}
We describe Quizz, a gamified crowdsourcing system that simultaneously assesses 
the knowledge of users and acquires new knowledge from them. Quizz operates by 
asking users to complete short quizzes on specific topics; as a user answers 
the quiz questions, Quizz estimates the user's competence. To acquire new 
knowledge, Quizz also incorporates questions for which we do not have a known 
answer; the answers given by competent users provide useful signals for
selecting the correct answers for these questions. Quizz actively tries to 
identify knowledgeable users on the Internet by running advertising campaigns, 
effectively leveraging  the targeting capabilities of existing, 
publicly available, ad placement services. Quizz quantifies the contributions 
of the users using information theory and sends feedback to the advertising
system about each user. The feedback allows the ad targeting mechanism to 
further optimize ad placement.

Our experiments, which involve over ten thousand users, confirm that we can 
crowdsource knowledge curation for niche and specialized topics, as the 
advertising network can automatically identify users with the desired expertise 
and interest in the given topic. We present controlled experiments that examine 
the effect of various incentive mechanisms, highlighting the need for having 
short-term rewards as goals, which incentivize the users to contribute. 
Finally, our cost-quality analysis indicates that the cost of our approach is 
below that of hiring workers through paid-crowdsourcing platforms, while 
offering the additional advantage of giving access to billions of potential 
users all over the planet, and being able to reach users with specialized 
expertise that is not typically available through existing labor marketplaces.
\end{abstract}

\section{Introduction}

Crowdsourcing has been the primary enabling mechanism behind the generation of 
many valuable Internet resources. Wikipedia, Freebase, and other knowledge 
repositories were created by volunteers who contributed knowledge about a wide 
variety of topics. Other human computation applications engage users in 
creative ways to generate interesting and useful by-products of the engagement. 
The ESP Game~\cite{von2004labeling} asks users to participate in a game that 
generates useful image tags. ReCAPTCHA~\cite{von2008recaptcha} verifies that 
users are humans by asking them to transcribe letters from a distorted image, 
thus helping with the digitization of books. 
Duolingo\footnote{\url{http://www.duolingo.com/}} teaches users a new language 
and as a by-product generates translations of written material in different 
languages. However, despite these widely-known success stories, building and 
engaging a community of users is a challenging task. Coming up with creative 
engagement strategies (e.g., ESP Game) is difficult, and spawning successful 
crowd-powered sites such as Wikipedia often seems like the exception rather 
than the norm.

In order to sidestep the problem, many efforts rely on paid crowdsourcing; for
example, hiring workers through platforms such as Amazon Mechanical Turk allows 
for direct engagement of users, with a clear monetary incentive. Unfortunately,
the introduction of money as a predictable and repeatable motivator is a mixed 
blessing. Users who are motivated by monetary rewards are often different than 
the users who are unpaid and motivated by other 
means~\cite{kuznetsov2006motivations,nov2007motivates,yang2010motivations}. 
Furthermore, studies indicate that \emph{the use of monetary rewards can be 
highly detrimental for users who are already intrinsically 
motivated}~\cite{ariely2009predictably}: the introduction of monetary 
compensation reveals to the users exactly how much their work is valued by the 
task requester, and low payments make things worse~\cite{gneezy2000pay}.

Finally, even if the incentive problem is solved, how does one attract the 
crowd that is properly qualified for a given task? The workers participating in
paid crowdsourcing are typically non-expert users, and often lack the skills 
needed for the crowdsourcing effort. For example, if one's task calls for 
Swahili speakers or maxillofacial surgeons, then most labor marketplaces do not
even provide access to such users, or only have few users with required expertise.

Thus, a set of natural challenging questions emerge. Can we replicate the 
predictability of paid crowdsourcing in terms of attracting participation, 
while engaging \emph{unpaid} users? And how can we identify and incentivize 
\emph{experts} among these users, who match the needs of the application at hand?

Here we propose to use existing Internet advertising platforms for targeting 
and attracting users, \emph{with the suitable expertise for the task at hand}. 
Over the last decade, advertising platforms have improved their targeting 
capabilities to identify users who are good matches for the goals of the 
advertiser. In our case, we initiate the process with simple advertising 
campaigns but also integrate the ad campaign with the crowdsourcing 
application, and provide \emph{feedback} to the advertising system for each  ad
click: The feedback indicates whether the user, who clicked on the ad, 
``converted'' and the total contributions of the crowdsourcing effort. This
allows the advertising platform to naturally identify websites with user 
communities that are good matches for the given task. For example, in our 
experiments with acquiring medical knowledge, we initially believed that 
``regular'' Internet users would not have the necessary expertise. However, the
advertising system automatically identified sites such as Mayo Clinic and 
HealthLine, which are frequented by knowledgeable consumers of health 
information who ended up contributing significant amounts of high-quality 
medical knowledge. Our idea is inspired by Hoffman et 
al.~\cite{hoffmann2009amplifying}, who used advertising to attract users to a 
Wikipedia-editing experiment, although they did not attempt to target users nor 
attempted to optimize the ad campaign by providing feedback to the advertising 
platform.

\begin{figure}[t]
\centering
\includegraphics[width=0.8\columnwidth]{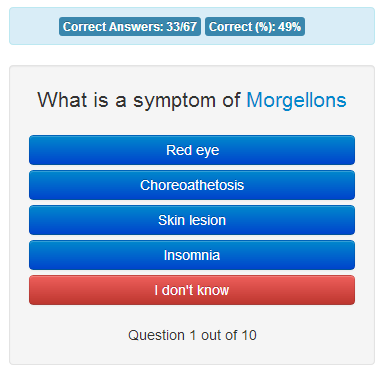}
\caption{Screenshot of the Quizz system.}\label{fig:quizz-screenshot} 
\end{figure}

Once users arrive at our site, we need to engage them to contribute useful 
information. Our crowdsourcing platform, \emph{Quizz}, invites users to test 
their knowledge in a variety of domains and see how they fare against other 
users. Figure~\ref{fig:quizz-screenshot} shows an example question. Our quizzes 
include two kinds of questions: \emph{Calibration} questions have known 
answers, and are used to assess the expertise and reliability of the users. On 
the other hand, \emph{collection} questions have no known answers and actually 
serve to collect new information, and our platform identifies the correct 
answers based on the answers provided by the (competent) participants. To 
optimize how often to test the user, and how often to present a question with 
an unknown answer, we use a Markov Decision Process~\cite{puterman2009markov}, 
which formalizes the exploration/exploitation framework and selects the optimal 
strategy at each point.

As our analysis shows, a key component for the success of the crowdsourcing 
effort is not just getting users to participate, but also to keep the good 
users participating for long, while gently discouraging low-quality users from 
participating. In a series of controlled experiments, involving tens of 
thousands of users, we show that a key advantage of attracting unpaid users 
through advertising is the strong self-selection of high-quality users to 
continue contributing, while low-quality users self-select to drop out. 
Furthermore, our experimental comparison with paid crowdsourcing (both paid 
hourly and paid piecemeal) shows that our approach dominates paid crowdsourcing 
both in terms of the quality of users \emph{and} in terms of the total monetary 
cost required to complete the task.

The contributions of this paper are fourfold. First, we formulate the notion of 
\emph{targeted crowdsourcing}, which allows one to identify crowds of users 
with desired expertise. We then describe a practical approach to find such 
users at scale by leveraging existing advertising systems. Second, we show how
to optimally ask questions to the users, to leverage their knowledge. Third, we
evaluate the utility of a host of different engagement mechanisms, which 
incentivize users to contribute more high-quality answers via the introduction 
of short-term goals and rewards. Finally, our empirical results confirm that 
the proposed approach allows to collect and curate knowledge with accuracy that 
is superior to that of paid crowdsourcing mechanisms at the same or lower cost.

\begin{figure}[t]
\centering
\includegraphics[width=\columnwidth]{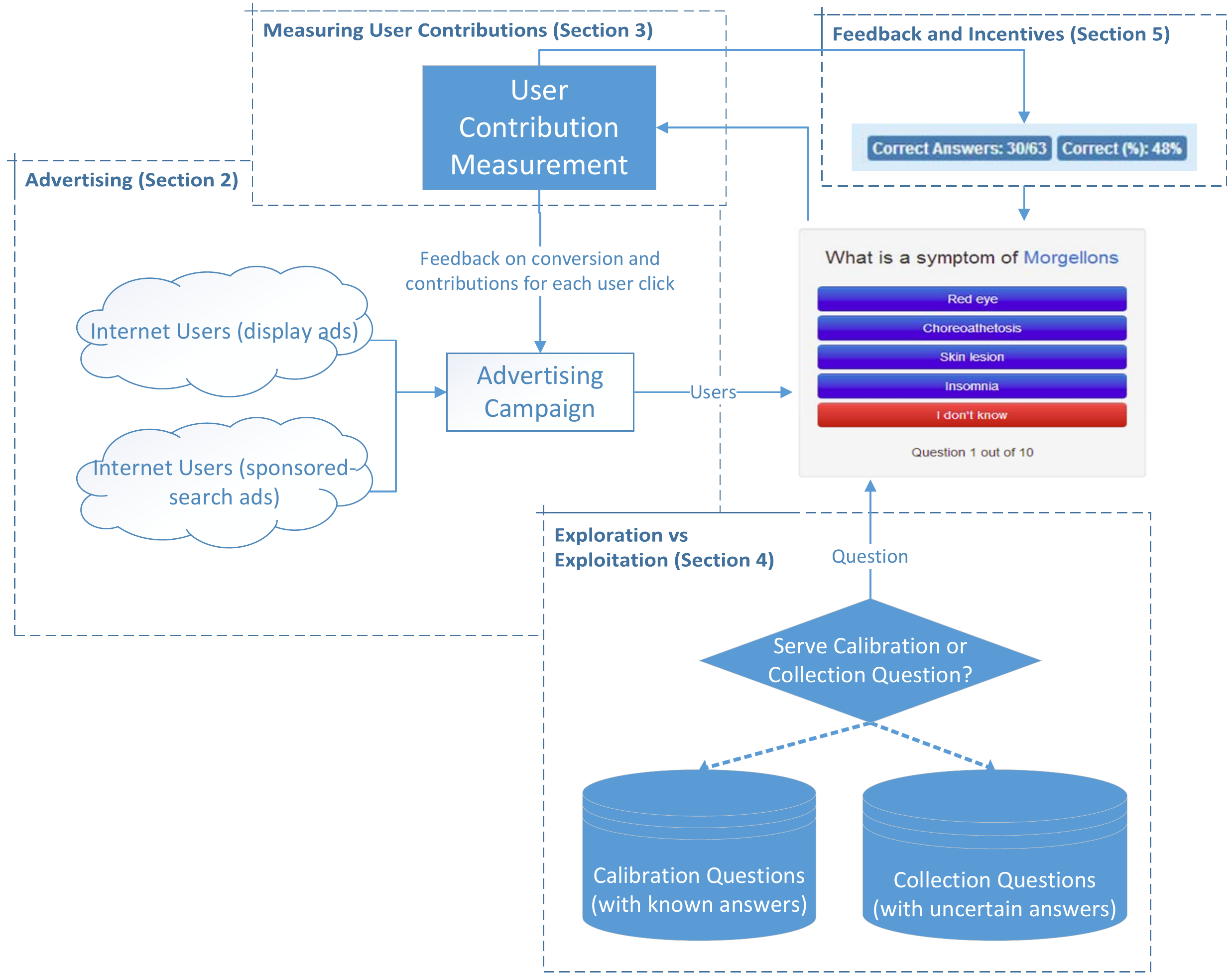}
\caption{An overview of the Quizz system.}\label{fig:quizz-overview}
\end{figure}

Figure~\ref{fig:quizz-overview} shows the overview of the system, and the 
various components that we discuss in the paper. Section~\ref{sec:advertising} 
describes the use of advertising to target promising users, and how we set up 
the campaigns to allow for continuous, automatic optimization of the results 
over time. Section~\ref{sec:bayesian-ig} shows the details of our 
information-theoretic scheme for measuring the expertise of the participants, 
while Section~\ref{sec:explore-exploit} gives the details of our 
exploration-exploitation scheme. Section~\ref{sec:incentives} discusses our 
experiments on how to keep users engaged, and Section~\ref{sec:experiments} 
gives the details of our experimental results. Finally, 
Section~\ref{sec:related} describes related work, while 
Section~\ref{sec:conclusions} concludes.

\section{Advertising for Targeting Users}
\label{sec:advertising}

A key problem of every crowdsourcing effort is soliciting users to participate. 
At a fundamental level, it is always preferable to attract users that have an 
inherent motivation for participation. Unfortunately, repeating the successes 
of efforts such as Wikipedia, TripAdvisor, and Yelp seems more of an art than  a
science, and we do not yet fully understand how to create engaging and viral 
crowdsourcing applications in a replicable manner. The emergence of paid 
crowdsourcing (e.g., Amazon Mechanical Turk) allows direct engagement of users 
in exchange for monetary rewards. However, the population of users who 
participate due to extrinsic rewards is typically different from the users who 
participate because of their intrinsic motivation.

\emph{Quizz} uses online advertising to attract \emph{unpaid} users to
contribute. By running ads, we get into the middle ground between paid and
unpaid crowdsourcing. Users who arrive at our site through an ad are not getting
paid, and if they choose to participate they obviously do so because  of their
intrinsic motivation. This removes some of the wrong incentives and tends to
alleviate concerns about indifferent users that ``spam'' the results just to get
paid, or about workers that are trying to do the minimum work necessary in order
to get paid. Thanks to the sheer reach of modern advertising platforms, the
population of unpaid users can potentially be orders of magnitude larger than
that in paid marketplaces. There are billions of users reachable through
advertising, while even the biggest crowdsourcing platforms have at most a
million users, many of them  inactive~\cite{ipeirotis2010demographics,
horton2010online}. Therefore, if the need arises (and subject to budgetary
constraints), our approach can elastically scale up to reach almost arbitrarily
large populations of users, by simply increasing the budget allocated to the
advertising campaign. At the same time, we show in Section~\ref{sec:experiments}
that our approach allows efficient use of the advertising budget  (which is our
only expenditure), and our overall costs are the same or lower than those  in
paid crowdsourcing installations.

\begin{figure}[t]
\centering
\fbox{\includegraphics[width=0.6\columnwidth]{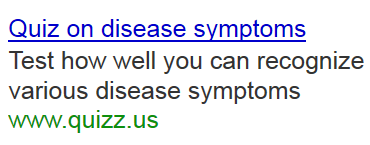}}
\caption{Example ad to attract users %to the Quizz application. 
%For this ad, we target the keyword ``disease symptoms.''
}\label{fig:quizz-ad}
\end{figure}

A significant additional benefit of using an advertising system is its ability
to \emph{target} users with expertise in specific topics. For example, if we are
looking for users possessing medical knowledge, we can run a simple ad like the
one in Figure~\ref{fig:quizz-ad}. To do so, we select keywords that describe the
topic of interest and ask the advertising platform to place the ad in relevant
contexts. In this study, we used Google
AdWords\footnote{\url{https://adwords.google.com}}, and opted into \emph{both
search and display ads}, while in principle we can use any other publicly
available advertising system.

Selecting appropriate keywords for an ad campaign is a challenging topic in
itself~\cite{fuxman2008using, abhishek2007keyword, joshi2006keyword}. However,
we believe that trying to optimize the campaign only through manually
fine-tuning its keywords is of limited utility. Instead, we propose to
automatically optimize the campaign by quantifying the behavior of the users
that clicked on the ad. A user who clicks on the ad but does not participate in
the crowdsourcing application is effectively ``wasting'' our advertising budget;
using the advertising terminology, such user has not ``converted.'' Since we are
not just interested in attracting any users but are interested in attracting
users who contribute, we use Google
Analytics\footnote{\url{http://www.google.com/analytics}} to track user
conversions. Every time a user clicks on the ad and then participates in a quiz,
we record a conversion event, and send this signal back to the advertising
system. This way, we are effectively asking the system to optimize the
advertising campaign for maximizing the number of conversions and thus
increasing our contribution yield, instead of the default optimization for the
number of clicks.

Although optimizing for conversions is useful, it is even better to attract
\emph{competent} users (as opposed to, say, users who just go through the quiz
without being knowledgeable about the topic). That is, we want to identify users
who are both willing to participate \emph{and} possess the relevant knowledge.
In order to give this refined type of feedback to the advertising system, we
need to measure both the quantity and the quality of user contributions, and for
each conversion event report the true ``value'' of the conversion. To achieve
this aim, we set up Google Analytics to treat our site as an e-commerce website, and for each conversion we
also report its  value. Section~\ref{sec:bayesian-ig} describes in detail our
approach to quantifying the values of conversions.

When the advertising system receives fine-grained feedback about conversions and
their value, it can improve the ad placement and display the ad to users who are
more likely to participate and contribute high quality answers. (In our
experiments, in Section~\ref{sec:experiments}, this optimization led to an
increase in conversion rate from 20\% to over 50\%, within a period of one
month, for a campaign that was already well-optimized.) For example, consider
medical quizzes. We initially believed that identifying users with medical
expertise who are willing to participate in our system would be an impossible
task. However, thanks to tracking conversions and modeling the value of user
contributions, AdWords started displaying our ad on websites such as Mayo Clicic
and HealthLine. These websites are not frequented by medical professionals but
by \emph{prosumers}. These users are both competent and are much more likely
than  professionals to participate in a quiz that assesses their medical
knowledge---often, this is exactly the type of users that a crowdsourcing
application is looking for.

\section{Measuring User Contributions}
\label{sec:bayesian-ig}

In order to understand the contributions of a user for each quiz, we need first
to define a measurement strategy.  Measuring the user contribution using just the number of
answers is problematic, as it does not consider the quality of the
submissions. Similarly, if we just measure the quality of the submitted answers,
we do not incentivize participation. Intuitively, we want users to
contribute high quality answers, and also contribute many answers. Thus, we need
a metric that increases as both quality and volume increase.

\textbf{Information Gain:} To combine both quality and quantity into a single,
principled metric, we adopt an information-theoretic
approach~\cite{waterhouse2013pay,robertson2009rethinking}. We
treat each user as a ``noisy channel,'' and measure the total information ``transmitted'' by the user during her
participation. The information is measured as the \emph{information gain}
contributed for each answer, multiplied by the total number of answers submitted
by the user; this is the total information submitted by the user. More formally,
assume that we know the probability $q$ that the user answers correctly a
randomly chosen question of the quiz. Then, the information gain $IG(q,n)$ is
defined as:
\begin{equation}
IG(q,n) = H(1/n, n) - H(q,n)
\end{equation}

\noindent where $n$ is the number of multiple choices in a quiz question. We use
$H(q,n)$ to define the entropy\footnote{Note that the user can select among $n$
possible answers, and we assume that the error probabilities are uniformly
distributed among the $n-1$ incorrect answers, each being selected with
probability $\frac{1-q}{n-1}$.} for an answer:
\begin{eqnarray}
H(q,n) & = & - \left( q \cdot \log(q) + \sum_{i=1}^{n-1} \left(\frac{1-q}{n-1}\right)\cdot \log\left(\frac{1-q}{n-1}\right)\right) \nonumber \\ 
	   & = & - q \cdot \log(q) - (1-q) \cdot \log\left(\frac{1-q}{n-1}\right)
\end{eqnarray}

\noindent When $q=1$ (user always gives perfect answers), then $H(q,n)=0$ (i.e.,
no uncertainty), and if $q=1/n$ (user selects randomly from the $n$ possible
answers)  then $H(q,n) = \log(n)$.

\textbf{Information Gain under Uncertainty:} In our environment, the quality $q$
of a user is unknown. In fact, the goal of Quizz is to \emph{estimate} $q$ for
each user, by asking the users to answer a set of quiz questions. We can try to
approximate $q$ with the ratio $q=\frac{a}{a+b}$, where $a$ is the number of
correct and $b$ is the number of incorrect answers for the user, but we face the
problem of sparse data, especially during the early stages of the quiz when
$a+b$ is relatively small.

Due to the uncertainty in measuring the exact quality of each user, we introduce
a Bayesian version of the information gain metric. Specifically, we explicitly
acknowledge the uncertainty of our measurements, and  we treat the estimate of
$q$  as a distribution, and not as a point estimate. The expected information
gain when $q$ is a random variable, we have:
\begin{equation}
\label{eq:big}
E[IG(q,n)] = \int_{q=0}^{1} Pr(q) \cdot IG(q,n)\,dq
\end{equation}

\begin{figure}[t] \centering
\includegraphics[width=0.8\columnwidth]{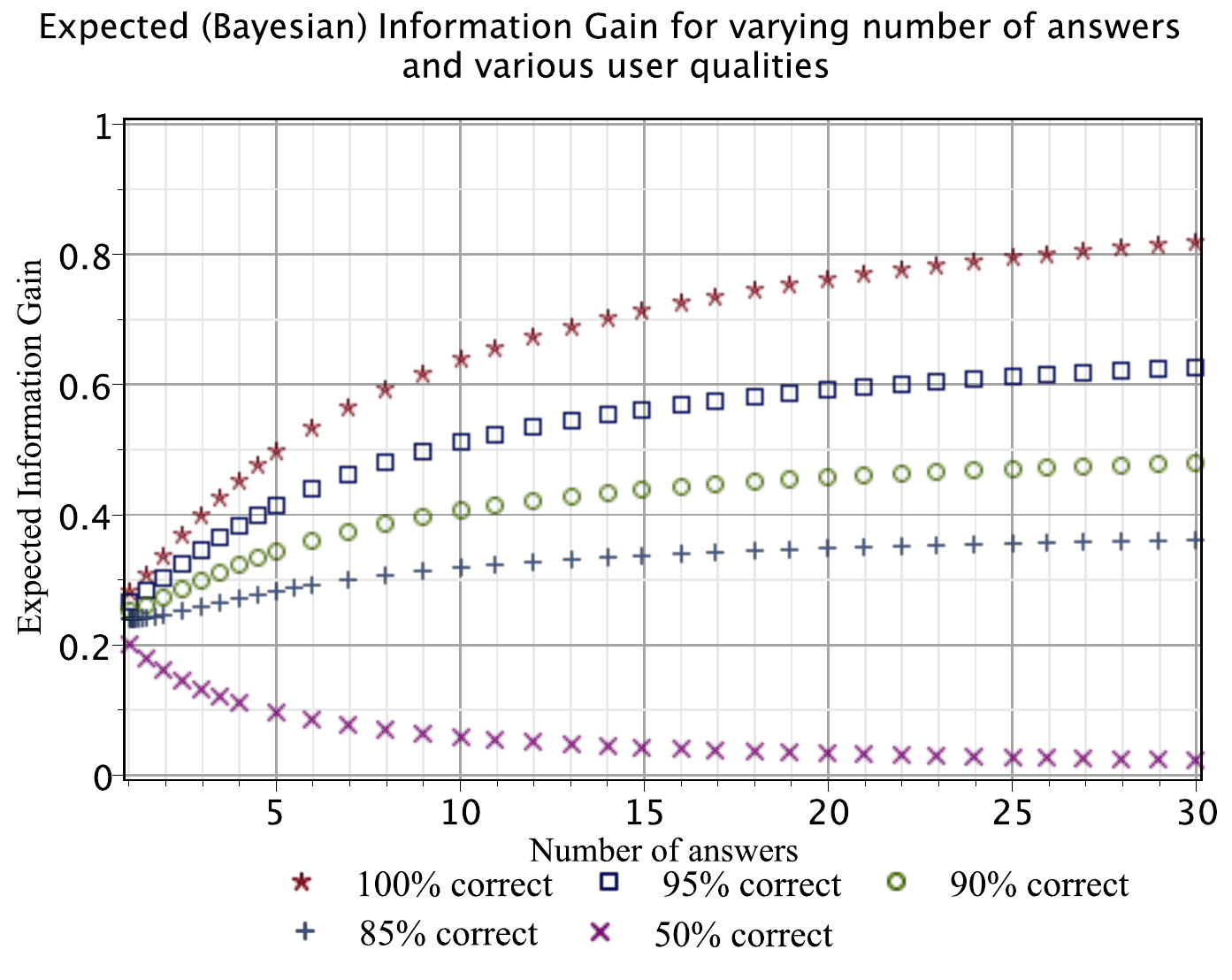}
\caption{The expected (Bayesian) information gain values, for varying number 
answers, and various user qualities, when the number of available answers for
each question $n=2$.}\label{fig:bayesIG}
\end{figure}

In our system, we assume that $q$ is constant across questions and
latent.\footnote{In the future, we can use Item Response
Theory~\cite{embretson2000item} and allow each question to have its own $q$
value.} However, we observe the number of correct answers $a$ from the user;
when $q$ is constant, $a$ follows a binomial distribution. We use the vanilla
Bayesian estimation strategy~\cite{gelman2003bayesian} for estimating the
probability of success $q$ in a binomial distribution. We set
$\mathit{Beta}(1,1)$ (i.e., the uniform distribution), as the conjugate prior
and then $Pr(q)$ is a Beta distribution.\footnote{Alternatively, we can use
a mixture of Beta priors to encode better our prior knowledge about the
distribution of user qualities~\cite{murphy2012machine}. } After the user
submits $a$ correct and $b$ incorrect answers, $Pr(q)$ follows the $\mathit{Beta}(a+1,b+1)$ distribution:
\begin{equation}
Pr(q) = q^a \cdot (1-q)^b \frac{1}{B(a+1, b+1)}
\end{equation}

\noindent with $B(x,y)$ being the Beta function. After some algebraic
manipulations, we have:
\begin{align}
E[IG(a,b,n)] = & \log(n) - \frac{b}{a+b} \cdot \log(n-1) - \Psi(a+b+1) \nonumber\\
 & + \frac{a \Psi(a+1)+b \Psi(b+1)}{a+b}
\end{align}

\noindent where $\Psi(x)$ is the digamma function. Figure~\ref{fig:bayesIG}
shows how $E[IG(a,b,n)]$ changes for different number of answers and for workers
of varying competence. Following the same process, we also compute the variance
of the information gain:
\begin{displaymath}
Var[IG(q,n)] = \int_{q=0}^{1} \left(IG(q,n)\right) ^2 \cdot Pr(q) \,dq - \left(E[IG(q,n)]\right)^2
\end{displaymath}

\noindent In the Appendix we list the detailed form of  $Var[IG(q,n)]$. In the
next section, we discuss how we use these measurements to optimally decide
between assessing the user's knowledge and collecting new judgments.

\section{Exploration / Exploitation}
\label{sec:explore-exploit}

\begin{algorithm*}[t]
\small
 \DontPrintSemicolon
 \KwData{Correct answers $a$, Incorrect answers $b$, Unknown answers $c$, Question utility $U^\mathit{past}_\mathit{question}$, Horizon limit $l$}
 \KwResult{Utility for all actions, Optimal next action}
 \Begin{

	\If{$l<0$} {
		return 0 \tcp*[l]{Reached the limit of computing horizon}
	}

	\BlankLine
	\BlankLine
	$\gamma = Pr(\mathit{survive}|\langle a, b, c\rangle)$
	\tcp*[l]{The (conditional) probability that the user will answer the served question.}

	$eig = E[IG(a,b,n)]$
	\tcp*[l]{Expected information gain}

	$sig = \sqrt{Var[IG(a,b,n)]}$
	\tcp*[l]{Standard deviation of information gain}
	
	$U^\mathit{now}_\mathit{question} = eig-sig$
	\tcp*[l]{The estimate of information gain for a question at the $\langle a, b, c \rangle$ state}

	\BlankLine
	\BlankLine
	
	\tcc{Utility estimation for a collection question}
	
	$U^\mathit{now}_\mathit{coll} = U^\mathit{now}_\mathit{question}$
	\tcp*[l]{If we ask a collection question, we get $eig-sig$ extra utility}

	$U^\mathit{future}_\mathit{coll} = \mathit{ComputeUtility}(a,b,c+1, U^\mathit{now}_\mathit{question}, l-1)$
	\tcp*[l]{Utility from future steps}

	$U_\mathit{coll} = \gamma \cdot (U^\mathit{now}_\mathit{coll} + U^\mathit{future}_\mathit{coll})$
	\tcp*[l]{Total utility of asking a collection question}

	\BlankLine
	\BlankLine
	
	\tcc{Utility estimation for a calibration question}
	
	$q = (a+1)/(a+b+n)$
 	\tcp*[l]{Probability of user answering correctly a calibration question}
 	 
 	 \BlankLine
 	\tcc{Utility if the user answers correctly} 	
	
	$U^\mathit{now/corr}_\mathit{question} = E[IG(a+1,b,n)]-\sqrt{Var[IG(a+1,b,n)]}$
	\tcp*[l]{Information gain, after a correct answer}
	
		$U^\mathit{now/corr}_\mathit{calib} = c \cdot ( U^\mathit{now/corr}_\mathit{question} - U^\mathit{past}_\mathit{question})$
	\tcp*[l]{Revise information gain for all $c$ previously-asked \emph{collection} questions.}
	
	$U^\mathit{fut/corr}_\mathit{calib} = \mathit{ComputeUtility}(a+1,b,c, U^\mathit{now/corr}_\mathit{question} , l-1)$
	\tcp*[l]{Utility from future steps, after a correct answer}

\BlankLine
	\tcc{Utility if the user answers incorrectly} 	
	 
	 $U^\mathit{now/incorr}_\mathit{question} = E[IG(a,b+1,n)]-\sqrt{Var[IG(a,b+1,n)]}$
	\tcp*[l]{Information gain, after an incorrect answer}
	
	$U^\mathit{now/incorr}_\mathit{calib} = c \cdot ( U^\mathit{now/incorr}_\mathit{question} - U^\mathit{past}_\mathit{question})$
	\tcp*[l]{Revise information gain for all $c$ previously-asked \emph{collection} questions.}
	
	 $U^\mathit{fut/incorr}_\mathit{calib} = \mathit{ComputeUtility}(a,b+1,c, U^\mathit{now/incorr}_\mathit{question}, l-1)$
	 \tcp*[l]{Utility from future steps, after an incorrect answer}
	
		\BlankLine
	$U_\mathit{calib} = \gamma \cdot (q \cdot (U^\mathit{fut/corr}_\mathit{calib} + U^\mathit{now/corr}_\mathit{calib}) + (1-q) \cdot (U^\mathit{fut/incorr}_\mathit{calib} + U^\mathit{now/incorr}_\mathit{calib}))$
	\tcp*[l]{Total utility of calibration}

  \BlankLine
   \BlankLine
  \eIf{$U_\mathit{calib} > U_\mathit{coll}$}{
  	Action = Ask calibration question\;
  }{
  	Action = Ask collection question\;
  }
  \Return{$\{U_\mathit{calib}, U_\mathit{coll}, U^\mathit{now}_\mathit{question}\}$, Action}\;

  }

 \caption{ComputeUtility($a$, $b$, $c$, $U^\mathit{past}_\mathit{question}$, $l$)}
 \label{algo:explore-exploit}
\end{algorithm*}

So far, we have described the setting where the user arrives and starts
participating by answering quiz questions. Using the information gain metric,
described in the previous section, we can estimate the amount of information
that we can extract from a user if we ask a \emph{collection} question, with an
unknown (to us) answer. However, our goal is not just to estimate how much
information we \emph{could} get, but actually acquire new knowledge from the
user. This creates a natural exploration-exploitation tradeoff. We can choose to
``explore'' how competent is the user, asking \emph{calibration} questions,
getting increasingly higher confidence about the user's competence on a topic;
or we can try to ``exploit,'' asking \emph{collection} questions.

To formalize our decision making, we assume that the decision on whether to
explore or exploit depends only on the current quiz that the user is solving and
the current state of the user, which can be described by the number of correct
answers $a$, the number of incorrect answers $b$, and the number of times $c$
that we asked a collection question. Given the state vector $\langle a, b,
c\rangle$ of the user, we use a Markov Decision Process (MDP) to select the next
action to take, based on the following considerations.

\begin{itemize}

\item \textbf{User dropping out:} At any point, the user may opt to abandon the
application. When the users drops out, we do not obtain any additional utility;
therefore an optimal set of actions should try to steer the user towards states 
with high probability of ``survival.'' (As we will see in 
Section~\ref{sec:experiments}, the probability of abandonment increases when 
the user gives incorrect answers to the quiz questions, and when the user does 
not receive feedback about the correctness of the submitted answer.) In our 
application, we estimate the probability $Pr(\mathit{survive}|\langle a, b,
c\rangle)$ based on the empirically-observed ``lifetimes'' of users, using a 
non-parametric kernel-density estimator with Gaussian  smoothing.\footnote{We
use the \texttt{KernSmooth} package in R.}  (See
Figure~\ref{fig:corr-incorr-distribution}.)

\item \textbf{Ask a collection question:} When we select to ask a collection
question, there are two components for the utility that the Quizz system
receives. Namely, there is immediate utility of getting information about the
potential answer for the question, and there is utility that we will accumulate
from the future actions of the user. The former utility is equal to the expected
information gain for the user given his current state vector $\langle a, b,
c\rangle$  (see Equation~\ref{eq:big}). However, we want to be more pessimistic
about the utility estimates and assign more value to learning about the user
competency. Following the ``value of learning'' approach~\cite{li2010value}, we
therefore set the reward to $U^\mathit{now}_\mathit{question} =
E[IG(q,n)]-\sqrt{Var[IG(q,n)]}$, in order to encourage our application to learn
more about the user before asking her to contribute new knowledge. The utility
for the future steps $U^\mathit{future}_\mathit{coll}$ is the utility for the
state vector $\langle a, b, c+1\rangle$, as we asked one more collection
question.

\item \textbf{Ask a calibration question:} When we select to ask a calibration
question, we  are trying to learn more about the competence of the user on the
specific topic of the quiz. When presented with a question, the user may give
either a correct or incorrect answer, which will lead to a revision of the
$E[IG(q,n)]$ and $Var[IG(q,n)]$ metrics. Although we do not get directly a
utility by asking a collection question, the revised estimate of
$U^\mathit{now}_\mathit{question}$ applies to all $c$ previously asked
collection questions. Therefore, the $U^\mathit{now}_\mathit{calib} = c \cdot
\left( U^\mathit{now}_\mathit{question} -
U^\mathit{past}_\mathit{question}\right)$ where
$U^\mathit{past}_\mathit{question}$ is the previous estimate of the utility of a
collection question. Furthermore, the utility of future steps, is the stochastic
sum of two possible forward paths: The utility
$U^\mathit{correct}_\mathit{calib}$ when the user gives the correct answer (with
probability $q=(a+1)/(a+b+n)$), and the utility
$U^\mathit{incorrect}_\mathit{calib}$ when the user gives an incorrect answer.

\end{itemize}

Algorithm~\ref{algo:explore-exploit} describes the implementation of this MDP.
One immediate concern with this formulation is that the recursive algorithm
definition points to states in the future, and hence the total reward could
potentially be infinite. This concern is alleviated if we assume that the
information gain in each step in bounded, and that the probability of survival
$\gamma = \max\{Pr(\mathit{survive}|\langle a, b, c\rangle)\} < 1$. We know that
the maximum information gain derived from a single question is $\log(n)$  and
there is always a non-zero probability that the user will abandon the
application. Therefore, the total utility that can be extracted from a single
user is bounded by  $\sum_i \gamma^i \cdot \log(n) \leq
\frac{\log(n)}{1-\gamma}$.

Another problem that arises with a recursive definition that points to future
states is that the computational estimation becomes harder. Classic dynamic
programming solutions assume a setting of backwards induction, where the
recursion eventually leads to some initial state that has a known utility (e.g.,
the recursive computation of $U(\langle a, b, c\rangle)$ depends on $U(\langle
a', b', c'\rangle)$ with $a'\leq a, b'\leq b, c'\leq c$). However, in our
setting we have a forward induction, making the computation of recursion
challenging. To allow the recursive computation to complete, we introduce a
limited execution horizon for the recursion~\cite{puterman2009markov}: once the
recursion has exceeded that level of depth, we stop the computation and return.
To find out whether the algorithm converges, we run the algorithm iteratively
with increasing horizon, until observing that the actions and utility
calculations converge.  Empirically, the algorithm converges faster when the
survival probabilities  $Pr(\mathit{survive}|\langle a, b, c\rangle)$ become
smaller.

\section{Engagement Incentives}
\label{sec:incentives}

In the previous section, we discussed how we can alternate between
``exploration'' and ``exploitation'' in order to assess the user's competence
and collect new information, respectively. A key requirement for the algorithm
to work effectively is to have a reasonable level of participation from the
users: if a user submits just a couple of answers, we cannot effectively assess
the user's competence or reliably collect new information.

Therefore, a key component of Quizz is the ability to continuously run
experiments with various incentive mechanisms, that are trying to incentivize
users to continue participating. Based on theories of intrinsic
motivation~\cite{malone1981toward}, we implemented a variety of incentives, with
the goal of prolonging the participation of competent users, while gently
discouraging the non-knowledgeable users from submitting low-quality answers.
Specifically, we tried the following options:

\begin{itemize}
 
  \item \textbf{Feedback for submitted answer:} We experimented with various
  types of feedback that we give back to the users. We tried giving \emph{no
  feedback}, \emph{saying whether the answer was correct or not}, and
  \emph{showing the correct answer}. The basic hypothesis is that immediate
  \emph{performance feedback}~\cite{malone1981toward} should motivate competent
  users to continue participating, and potentially incentivize low-performing
  users to try to improve their performance.
 
  \item \textbf{Displaying scores:} We experimented with displaying different
  types of scores to the user. We displayed the \emph{percentage of correct
  answers}, the \emph{total number of correct answers}, and a score based on the
  \emph{information gain}, and combinations thereof.
 
  \item \textbf{Displaying crowd performance:} We experimented with showing the
  \emph{crowd performance} on each question (i.e., how many users answered the
  question correctly). We hypothesized that knowing how other users perform is
  going to increase the user effort.
 
  \item \textbf{Leaderboards:} We experimented with showing the ranking of the
  user compared to other participants. Our hypothesis was that users will modify
  their behavior~\cite{anderson2013steering,easley2013incentives} striving to
  reach one of the top leaderboard positions.

\end{itemize}

\section{Evaluation}
\label{sec:experiments}

\subsection{Metrics}
\label{sec:metrics}

In order to evaluate our system, we used the following metrics to quantify the
level of user engagement and the quality of their contributions.

\begin{itemize}
  
\item \textbf{Conversion rate:} We define conversion rate as the fraction of
users who answered at least one quiz question, after clicking one of our ads
(cf.\ Section~\ref{sec:advertising}). We use this metric to measure the
effectiveness of our advertising.
   
\item \textbf{User lifetime:} We examine the number of (correct and incorrect)
answers submitted by the users. We use this metric mainly to understand the
effect of the various engagement incentives.
	
\item \textbf{Total information gain:} We measure the expected information gain
of each user using Equation~\ref{eq:big}, and multiply this value by the total
number of answers submitted by the user. The result is the total information
that we received from the user.
	
\item \textbf{Monetary cost per correct fact:} We measure the total cost
required to verify a fact at the 90\%, 95\%, and 99\% estimated accuracy. To
compute the cost for various levels of accuracy, we use the fact that if we know
the quality of a contributor, we can estimate the required redundancy to reach
the desired level of confidence~\cite{wang2013quality, genest1986combining}. For
example, if we have two users that are 90\% accurate and we pay a cost of \$0.10
per contributed answer, we need one such worker to verify a fact at 90\%
accuracy (i.e., cost \$0.10 at 90\% accuracy), and  approximately two such
workers to verify a fact at the 99\% (i.e., cost \$0.20 at 99\% accuracy). To evaluate
the correctness of the answers submitted by the users and the corresponding
capacity of the system, we used questions with answers that have been
pre-validated by multiple, trusted human judges. The costs are calculated based
on the total advertising expenditure for attracting the users to the Quizz site,
broken down by quiz.

\end{itemize}

\subsection{Capacity and cost analysis}
\label{sec:capacity}

\begin{table*}[t]
\footnotesize
\center
\begin{tabular}{lrrrcccccc}
\toprule
Quiz	&	Users	& Answers &	Cost	&	\multicolumn{3}{c}{Capacity/User} &	\multicolumn{3}{c}{Cost/Answer} \\
	    &		&		&	 &      @99\%	 &	 @95\%	&	 @90\%	&	@99\%	&	@95\%	&	@90\% \\
\midrule
Disease Causes	&	414	& 7,644	&	\$51.13	&	3.75	&	4.83	&	6.49	&	\$0.07	&	\$0.05	&	\$0.04\\
Disease Symptoms	&	569	& 11,088	&	\$12.51	&	3.30	&	4.25	&	5.71	&	\$0.02	&	\$0.01	&	\$0.01\\
Treatment Side Effects	&	605	& 5,044	&	\$46.38	&	1.22	&	1.57	&	2.12	&	\$0.13	&	\$0.10	&	\$0.07\\
Artist and Albums	&	310	& 1,548	&	\$21.56	&	0.88	&	1.13	&	1.52	&	\$0.16	&	\$0.13	&	\$0.09\\
Latest Album	&	522	& 2,588	&	\$20.70	&	0.95	&	1.23	&	1.65	&	\$0.09	&	\$0.07	&	\$0.05\\
Artist and Song	&	925	& 5,285	&  \$236.26	&	0.96	&	1.23	&	1.66	&	\$0.54	&	\$0.42	&	\$0.31\\
Film Directors	&	412	& 2,250	&	\$16.49	&	1.19	&	1.54	&	2.07	&	\$0.07	&	\$0.05	&	\$0.04\\
Movie Actors	&	337	& 2,189	&	\$36.14	&	0.96	&	1.24	&	1.66	&	\$0.22	&	\$0.18	&	\$0.13\\
\midrule
\em Average	&	\em 512	& \em 4,704	& \em \$55.15	& \em	1.65	& \em	2.13	& \em	2.86	& \em	\$0.16	& \em	\$0.12	& \em	\$0.09\\
\bottomrule
\end{tabular}
\caption{Answers that can be collected per user and corresponding cost at 90\%,
95\%, and 99\% accuracy levels. The capacity per user metric is the number of
collection questions that can be answered using the Quizz system, at different
levels of accuracy. Cost is the amortized cost per question. All campaigns run
with the conversion optimizer enabled, and a target cost-per-conversion of
\$0.10.}
\label{table:capacity-cost}
\end{table*}

Based on our measurements for September 2013, the conversion rate for the Quizz
application was an average of 34.60\%, resulting in a total of 4,091 engaged
users out of 11,825 users that visited the application (having clicked an ad).
The conversion rate increased steadily over time, starting at around 20\% in the
beginning of the month, and reaching a high of 51.25\% on September 30th. (As we
will discuss below, this is due to the continuous optimization from the
conversion optimizer). 

In our experiments we used eight different quizzes on
various topics, and we report in Table~\ref{table:capacity-cost} the detailed
results in terms of user recruitment, cost, and the capacity of the system in
providing answers to new questions at different accuracy levels. On average, we
paid \$0.16 to validate a fact at the 99\% accuracy level, and \$0.09 to
validate a fact at the 95\% accuracy level. An interesting
outlier is the ``Artist and Song'' quiz, which ended up having significantly
higher costs per collected answer compared to the other quizzes. In this case,
effectively we observed a failure of the advertising system to identify a group
of users that would be knowledgeable of \emph{all} the artists and songs that
appeared in the quiz: given the diversity of music genres in that quiz, it was
difficult to find music fans that have \emph{detailed} knowledge of all the
song titles of various artists across multiple genres. (However, it \emph{was}
possible to find music fans that have knowledge of at least the album names of 
different artists, an admittedly easier task.)

\subsection{User contributions and self-selection}
\label{sec:lifetime}

\begin{figure}[t]
\centering
\includegraphics[width=\columnwidth]{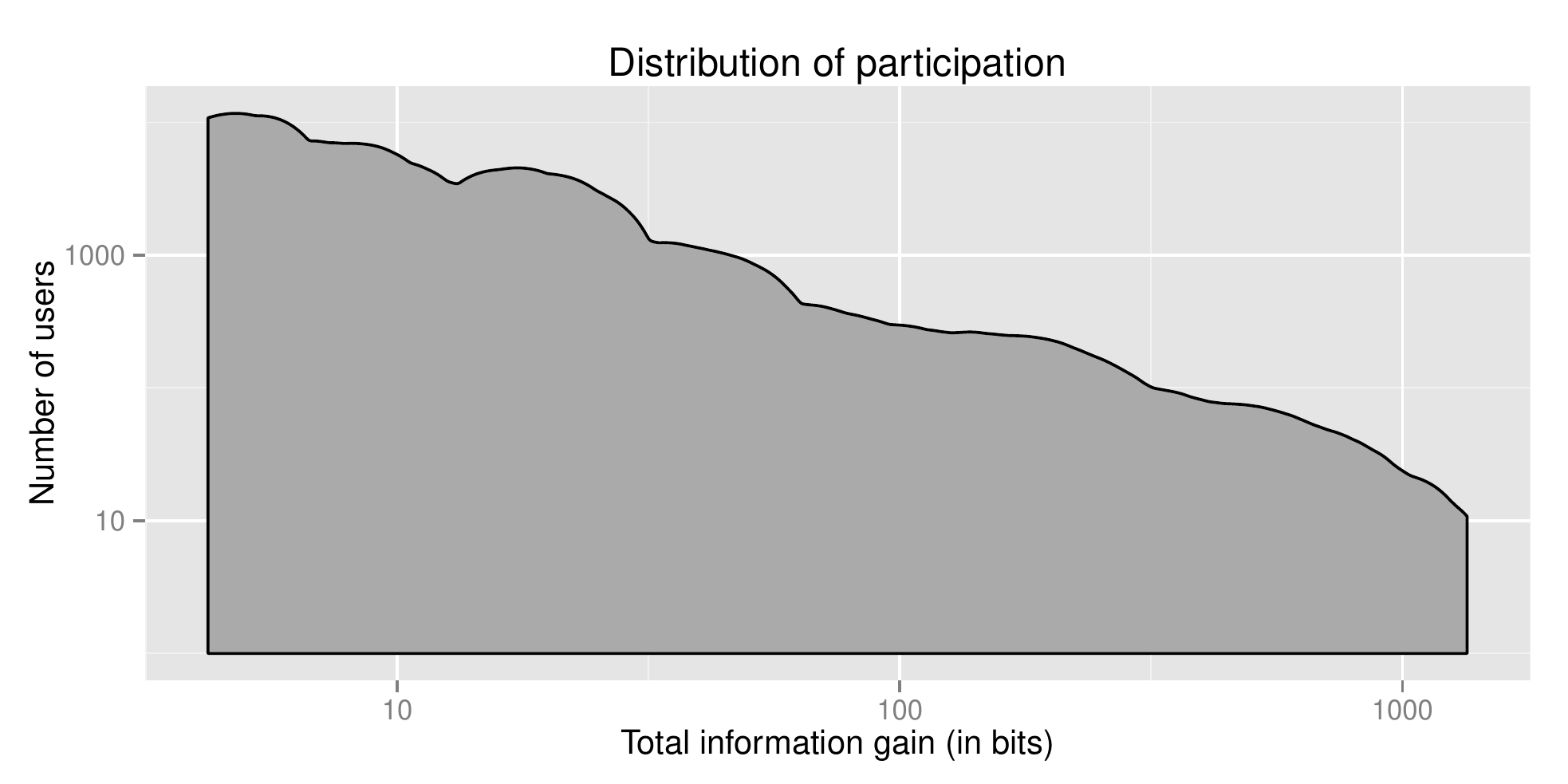}
\caption{Distribution of participation. The vertical axis is the the number of users (in log scale), the horizontal axis is the of the total information gain. }\label{fig:participation-distribution}
\end{figure}

\begin{figure}[t]
\centering
\includegraphics[width=\columnwidth]{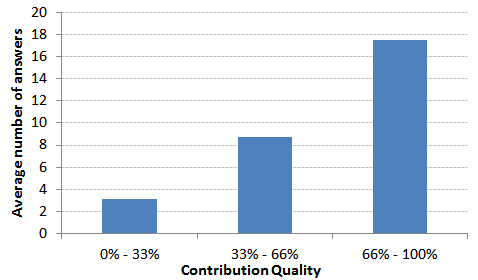}
\caption{Lifetime of users according to their contribution quality.}\label{fig:corr-incorr-distribution}
\end{figure}

\begin{figure}[t]
\centering
\includegraphics[width=\columnwidth]{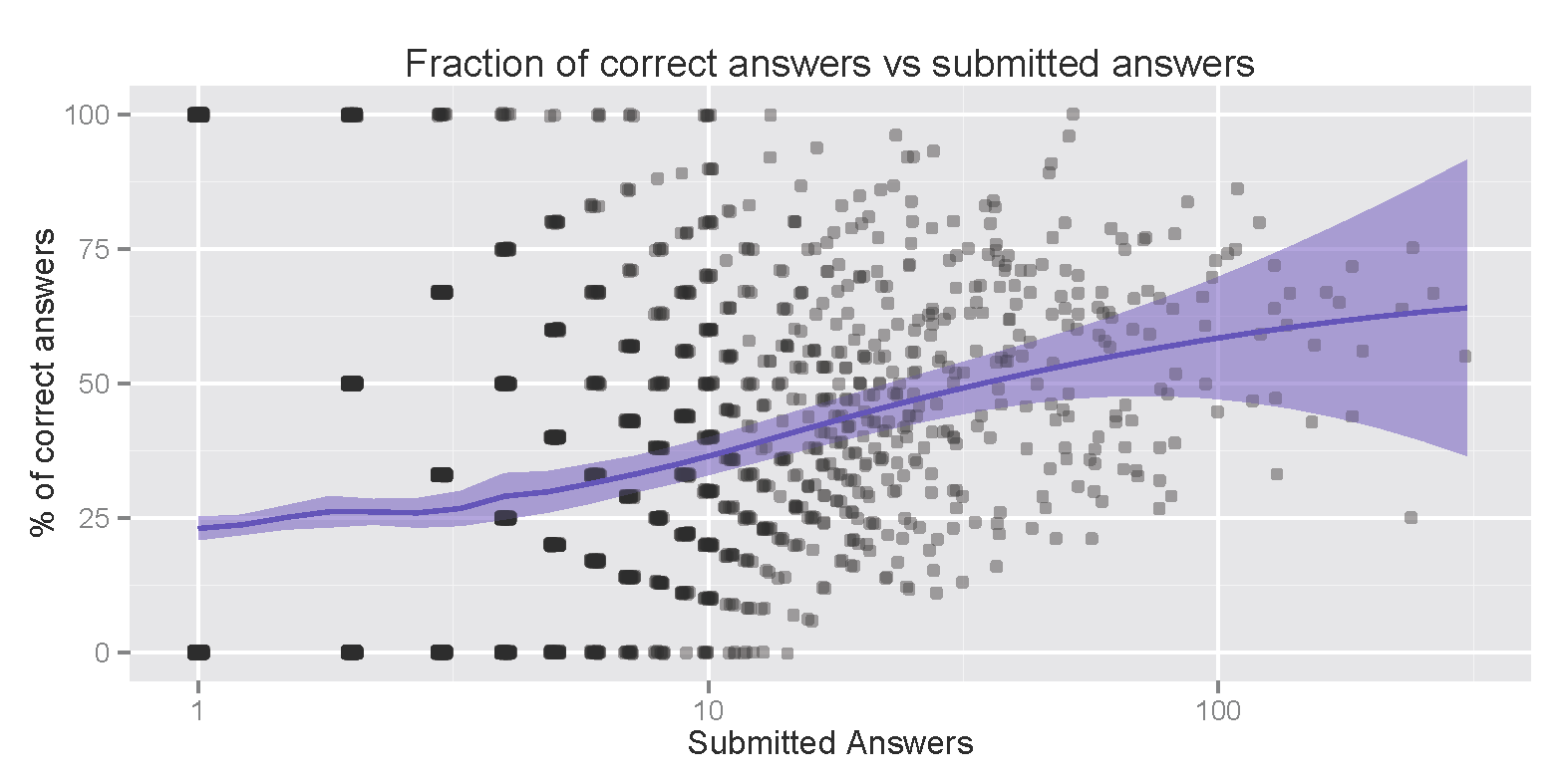}
\caption{Quality of submissions as a function of user participation. (Intensity
of dot color corresponds to the number of users represented by the dot; fitted
line computed using LOESS.) Knowledgeable users self-select and to continue
submitting answers, leading to significant increase of answer quality for 
heavy participants. Low-performers drop out voluntarily without submitting 
many answers.}\label{fig:quality-vs-participation}
\end{figure}

In terms of a per-user contribution, Figure~\ref{fig:participation-distribution}
shows the distribution of total information gain across the participating users,
and Figure~\ref{fig:corr-incorr-distribution} shows the lifetime of users as a
function of their quality. As expected, many users come, submit a few answers,
and then leave. These are the ``head'' users; although they do contribute some
useful signal, they do not generate a great ``return on investment.''
Figure~\ref{fig:quality-vs-participation} further illustrates that the users
that submit large number of answers also tend to submit more correct answers
than incorrect. This means that the users who are competent about the topic
submit more and more answers, while the ones who cannot answer the quiz
questions correctly, drop out. This is an illustration of the benefit of unpaid
users: there is little incentive for unpaid users to continue participating when
there is no monetary reward and they are not good at the task. (We present a
more detailed comparison with paid crowdsourcing in
Section~\ref{sec:comparison-paid}.)

\subsection{The effect of targeting in advertising}
\label{sec:targeting}

\begin{figure}[t] \centering
\includegraphics[width=\columnwidth]{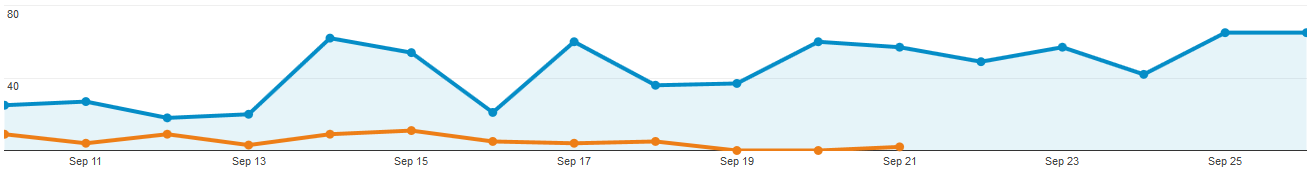}
\includegraphics[width=\columnwidth]{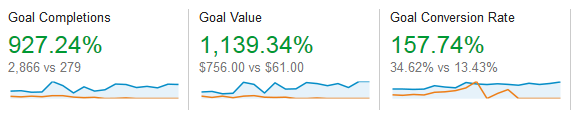} \caption{Comparison
of conversions for targeted vs.\ untargeted ad campaign (screenshot from Google
Analytics; the upper plot shows the information gain for the two campaigns
as a function of time; blue: targeted, yellow: untargeted). The
targeted campaign generates 9.2x more conversions, as well as higher-quality answers.}\label{fig:targeted-vs-untargeted}
\end{figure}

A major hypothesis of our work is that the targeting system of existing
advertising networks can be leveraged in order to identify competent users, who
are willing to contribute new knowledge by answering quiz questions. We 
observe that users recruited through advertising are knowledgable and willing to
contribute. However, it is not immediate obvious whether the positive result is
due to targeting, or is simply the effect of bringing more users to the
application.

In order to disentangle the effects of advertising and targeting, we ran two
different advertising campaigns that both directed users to the same quiz. Both
campaigns had the same budget, same ad creatives, same bidding settings, 
and their only differences were (1)~the keywords used for the
bidding and (2)~the use of feedback to the advertising system. In the targeted
 campaign, we used keywords related to the topic of the quiz; for the
untargeted campaign we used the keywords from all the quizzes available in the
Quizz system. Also, in the untargeted campaign, we did not send feedback about
conversions to avoid providing targeting information.

Figure~\ref{fig:targeted-vs-untargeted} shows the results. While the number of
visitors was roughly the same for the two campaigns, the targeted campaign had
3x higher conversion rate (34.62\% vs.\ 13.43\%). Furthermore, among the
participating (``converted'') users, the number of questions answered per user
was 3x higher for the users who arrived from the targeted campaign. Thus, the
cumulative difference was over 9.2x more answers obtained through the targeted
campaign compared to the untargeted one (2866 answers vs.\ 279). Finally, the
answers contributed by the users from the targeted campaign were of higher
quality than the answers from the untargeted campaign: the total information
gain for the targeted campaign was 11.4x higher than the total information gain
for the untargeted one (7560 bits vs.\ 610 bits), indicating a higher user
competence, even on a normalized, per-question basis.

\subsection{The effect of using conversion optimizer}
\label{sec:optimizer}

 After verifying that targeting and feedback indeed improve the results, we
 wanted to examine the effect of using the \emph{conversion optimizer}. While
 traditional ad campaigns usually optimize for clicks, the conversion optimizer
 of Google AdWords offers the option to optimize for the total ``value'' of the
 conversions (in our case, for the total information gain). To examine the
 usefulness of the conversion optimizer in our setting, we again run two
 otherwise-identical ad campaigns: one being optimized for clicks, and the
 other being optimized for conversions.

 The conversion rate increased by 30\% when using the conversion optimizer (from
 29\% to 39\%). In addition to that, the number of submitted answers went up by
 42\% (1683 vs.\ 1183), and the total information gain went up by 63\% (4690
 bits vs.\ 2870 bits). Furthermore, as Section~\ref{sec:capacity} discusses, the
 optimization is ongoing and the conversion rate continues to go up even at the
 time of this writing. This automatic and continuous optimization of the process
 illustrates the benefits of leveraging existing, publicly available advertising
 platforms to improve the efficiency of crowdsourcing applications.

\begin{table}[t] \footnotesize \center
\begin{tabular}{lrl}
\toprule
\bf Variable: Total	&	Coefficients	& 	Significance\\
\midrule
%(Intercept)	&	1.456835	&	***\\
showCorrect	&	0.142475	&	***\\
showMessage	&	-0.008423	&	\\
showPercentageCorrect	&	-0.003646	&	\\
showTotalCorrect	&	0.085231	&	***\\
showScore	&	0.093463	&	***\\
showCrowdAnswers	&	0.025502	&	*\\
showPercentageRank	&	-0.074008	&	***\\
showTotalCorrectRank	&	-0.006259	&	\\
\midrule	
\bf Variable: Correct	&		&	\\
\midrule
%(Intercept)	&	0.536113	&	***\\
showCorrect	&	0.13936	&	***\\
showMessage	&	0.002874	&	\\
showPercentageCorrect	&	0.010314	&	\\
showTotalCorrect	&	0.085838	&	***\\
showScore	&	0.062361	&	***\\
showCrowdAnswers	&	0.041233	&	*\\
showPercentageRank	&	-0.101775	&	***\\
showTotalCorrectRank	&	-0.007062	&	\\
\midrule	
\bf Variable: Score	&		&	\\
\midrule
%(Intercept)	&	5.016456	&	***\\
showCorrect	&	0.204104	&	***\\
showMessage	&	0.024464	&	**\\
showPercentageCorrect	&	0.023831	&	***\\
showTotalCorrect	&	-0.022065	&	***\\
showScore	&	0.040387	&	***\\
showCrowdAnswers	&	0.098557	&	***\\
showPercentageRank	&	-0.048658	&	***\\
showTotalCorrectRank	&	-0.016987	&	***\\
\bottomrule
\end{tabular}
\caption{The effect of various mechanisms in incentivizing users to submit  more
answers \emph{Total}, more correct answers \emph{Correct}, and to improve  their
total information gain \emph{Score}. The coefficients were computed by a 
Poisson regression model (***: 0.1\% significance, **: 1\% significance, *: 5\%
significance). Given that $e^x \approx 1+x$ for small values of $x$, a
 coefficient of 0.1 means that we observe a 10\% improvement, while a coefficient of -0.1 means that we observe a 10\%
decrease in performance.}
\label{table:poisson-incentives}
\end{table}

\subsection{The effect of engagement incentives}
\label{sec:results-incentives}

 To analyze the effect of the various incentive mechanisms, we examined how the
 different experimental conditions assigned to the users affected their
 participation and their overall contributions. To this end, we examined the
 effect of the various incentives on three variables of interest: the
 \emph{total} number of submitted answers, the number of \emph{correct} answers,
 and the (total information gain) \emph{score} of the user. Since the dependent
 variables are always positive and behave like ``count data'' we ran a
 Poisson regression, with eight binary variables as dependent variables, where
 each of these variables corresponded to the presence (or absence) of an
 experimental condition. Specifically, we present results for the following
 incentive mechanisms:

\begin{itemize}
\item \textbf{showCorrect}: Show the correct answer.

\item \textbf{showCrowdAnswers}: Show the percentage of other users who
answered the question correctly.

\item  \textbf{showMessage}: Show whether the given answer was correct.

\item \textbf{showPercentageCorrect}: Show the percentage of submitted
answers (for that user) that were correct.

\item 	\textbf{showTotalCorrect}: Show the total number of correct answers submitted by that user.

\item \textbf{showScore}: Show the total information gain for the user (shown as a score).

\item  \textbf{showPercentageRank}: Show the position of the user in the leaderboard, ranked by percentage of correct answers.

\item   \textbf{showTotalCorrectRank}: Show the position of the user in the leaderboard, ranked by the total number of correct answers submitted.
\end{itemize}

Table~\ref{table:poisson-incentives} summarizes the results, and shows the
coefficients computed for each mechanism by the regression model. Showing the
correct answer (showCorrect) has the strongest impact in increasing
participation, as it has strong positive effect across all three
dependent variables, indicating that users want to know what the correct answer is.
Interestingly, knowing whether they were correct or not (showMessage) does not
have a similarly strong effect. These results indicate that users may be more
interested in \emph{learning} about the topic rather than just knowing whether
they answered correctly.

Experimenting with the performance-related incentives (i.e.,
showPercentageCorrect, showTotalCorrect, showScore) generated some interesting
observations. Showing the percentage of correct answers did not have a
statistically significant effect in terms of answer counts, but had a slightly
positive effect in the total information gain. Showing the total number of
correct answers generated an interesting effect: while both total and correct
answers went up, the total information gain was affected negatively. It appears
that non-competent users were also positively influenced to participate more,
leading to a decrease of the overall answer quality. Not surprisingly, when we
show the total information gain as a score to the user, this effect disappears,
and we observe positive outcomes across all variables.

Finally, the competitive incentives (i.e., showCrowdAnswers, showPercentageRank,
showTotalCorrectRank) demonstrated an interesting behavior of the users: knowing
the performance of other users has a positive effect in the participation, which
indicates that users are interested in how they fare against other users.
However, displaying leaderboards had a generally negative effect across all
variables. Interestingly enough, if we examine the effect of leaderboards in the
early stages of the application, we see a very strong \emph{positive} effect in
terms of participation. Our hypothesis to explain these contradicting
observations is the following. Early on, the leaderboards are relatively
sparsely populated and it is relatively easy for users to go up and reach some
of the top positions. However, as more and more users participate, the
achievements of the top users are difficult to match, effectively discouraging
users from trying harder. To test our hypothesis, we ran a small experiment,
where the leaderboard was computed based on the participation from last week, as
opposed to showing an all-time leaderboard. The results indicated that the
``last-week'' leaderboard with fewer and less impressive achievements has indeed
had a \emph{positive} effect on participation, compared to the ``all-time''
leaderboard. This indicates that users are motivated by the potential for
achievement, and by showing the users that they can reach an achievement in a
relatively easy manner can help with participation.

% Perhaps lying? Perhaps daily winner?

\begin{figure}[t]
\centering
\includegraphics[width=\columnwidth]{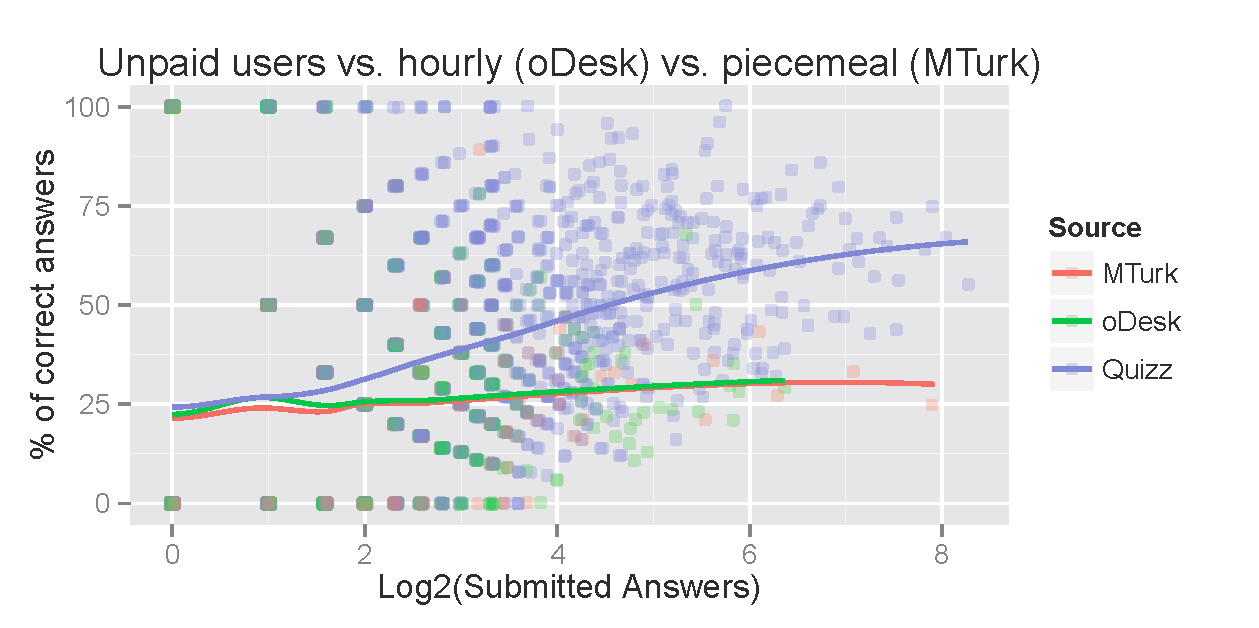}
\caption{Comparison of unpaid users vs.\ common payment schemes (hourly and piecemeal payments). Paid workers are not experts in the presented topic, and their quality is barely above random, despite the bonus incentives for good performance; furthermore, the monetary reward incentivizes low-perfoming paid workers to continue participating.}\label{fig:quizz-vs-paid}
\end{figure}

\subsection{Comparison with paid crowdsourcing}
\label{sec:comparison-paid}

Finally, we wanted to compare the performance of our approach against a pure
paid-crowdsourcing setting. To this end, we hired workers through Mechanical
Turk, and paid them 5 cents per question (i.e., piecemeal payment), with an
extra bonus that depended on their total score (information gain) at the end.
Similarly, we hired workers via oDesk, paid them on an hourly basis (ranging
from \$5/hr to \$15/hr, depending on their asking price), and we also indicated
that they will receive an additional payment based on their overall score.
Figure~\ref{fig:quizz-vs-paid} summarizes the results. Our key observation is
that the workers hired through paid crowdsourcing platforms are usually not
experts in the topic of the quiz, and are therefore not sufficiently
knowledgeable to provide high quality answers. However, unlike the unpaid
workers, the paid workers have an obvious monetary incentive to continue
working, and so we did not observe the self-selection dropout effect for the
paid workers. The paid workers continue submitting low-quality answers, and this
finding is similar with both piecemeal and hourly payments.

While there are some compeqtent workers among the paid participants, the total
information gain from the competent paid workers is still significantly lower
than the information gain from unpaid users, resulting in significantly lower
capacities. The \emph{best} paid worker had a 68\% quality for the quiz, and
submitted 40 answers, resulting in an equivalent capacity of 13 answers at 99\%
accuracy, or 23 answers at 90\% accuracy. To match the performance of the unpaid
users, the worker should be paid 5 cents per question, or \$3/hr, taking into
account that the average time per question \emph{for the paid users} is one
minute. (For comparison, unpaid users are much faster and typically give an
answer within 10 seconds, signaling that they are already knowledgeable about
the topic of hte quiz and they do not perform research to answer the questions.)
Given that all other workers demonstrated worse metrics, it is clear that
unpaid, volunteer users dominate. A potential solution is to experiment with
negative incentives (e.g., ``you will not get paid unless you achieve this
quality score''), keeping away the low perfomers, and keeping just the top
workers. However, it is not clear how we can reach these high-quality workers in
a labor market, other than by posting the task and then hoping that the
competent workers will participate. Potentially, labor marketplaces can employ
targeting schemes, similar to the ones we implemented using online advertising,
but today we are not aware of any marketplace offering such functionality.

\section{Related Work}
\label{sec:related}

Quizz crowdsources the acquisition of knowledge by asking users to participate
in thematically-focuses quizzes, which contain also ``collection'' questions
with no known answer. ReCAPTCHA~\cite{von2008recaptcha} is close conceptually,
as it asks users to type two digitized words, out of which one is known and the
other is unknown, which is similar to our calibration and collection questions,
respectively. 
In terms of use of advertising for recruiting users, Hoffman et
al.~\cite{hoffmann2009amplifying}  use advertising to attract participants for a
Wikipedia-editing experiment; however there was no discussion or experiments
with targeting, or with optimizing the ad campaigns for maximizing the user
contributions.

Recent work~\cite{anderson2013steering,easley2013incentives} built models on how
badges and leaderboards should be designed to engage users and steer their
behavior towards actions that are beneficial for the system. Our work
empirically tests some of these models, and our experimental results dovetail
the suggestions of these models. Other models of user engagement have examined
what metrics and measurments capture the user level of
engagement~\cite{lehmann2012models, attfield2011towards, baeza2012user,
dupret2013absence, mccay2012saliency}. Our analysis of engagement focuses mainly
on web analytics measurements, without trying to interact further
with the participating users, although this is a promising direction for future
work.

In our work, we explicitly assess the competence of users with calibration
questions. Alternatively, we can use unsupervised techniques for estimating the
competence of users, through redundancy. Dawid and Skene~\cite{applstat/dawid79}
presented an EM algorithm to estimate the quality of the participants in the
absence of known ground truth, and a large number of recent papers examined the
same
topic~\cite{raykar2010learning,whitehill2009whose,welinder2010multidimensional}
improving significantly the state of the art. Being closer to our work, Kamar et
al.~\cite{kamar2012combining} also use a Markov Decision Process, in order to
decide whether the answers provided by a user are promising enough to warrant a
hiring decision. In the future, we plan to use these algorithms for quality
inference together with our exploration/exploitation approach, to decide
optimally how to combine assessment with knowledge acquisition. A key challenge
is being able to provide immediate feedback to the users, when the
 questions have no certain answer.

Optimal acceptance sampling plans in quality control~\cite{dodge73, wetherill75,
berger82, schilling82} is another related line of work. The purpose of
acceptance sampling is to determine how much to sample a production line, in
order to decide whether to accept or reject a production lot. The key difference
with our setting is the limited lifetime of the users (as opposed to the
significantly higher production capacity in industrial production), and our
planning needs to be much more dynamic than in the most use cases of acceptance
sampling.

\section{Conclusions}
\label{sec:conclusions}

We presented a model for targeting and engaging \emph{unpaid} users in a
crowdsourcing application. We demonstrated how to use existing Internet
advertising platforms to identify niche audiences of competent users for the
task at hand, and we showed that using publicly available ad-optimization tools
can result in significant improvements in the effectiveness of the process.
Currently, our application has a 50\% conversion rate for every ad click, and
the cost per answer drops systematically over time, as the advertising system
learns to identify competent users that are likely to be high contributors. The
engagement of unpaid users alleviates concerns about the incentives of paid
users, who are not always well-aligned with the goals of the crowdsourcing
application. Furthermore, our algorithms and controlled real-life experiments
with over ten thousand users illustrate how to setup incentive mechanisms in
practice to engage users and extend their ``lifetime'' in the system. Finally,
our experiments indicate that even though there are costs associated with
advertising, the quality-adjusted costs are on par with those of paid
crowdsourcing. (Moreover, for non-profits, engaging for example in 
\emph{citizen science} efforts, there are ways to get a substantial advertising 
budget using programs such as ``\emph{Google Ad Grants for
nonprofits},''\footnote{\url{http://www.google.com/grants/}} which offers
\$10,000 per month in in-kind  advertising budget.)  We believe that our
ad-based approach can form the foundation towards more predictable deployment
and engagement of unpaid users in crowdsourcing efforts, combining the advantage
of engaging unpaid users with the predictability of paid crowdsourcing.

\section*{Acknowledgments}

We would like to thank Kevin Murphy and Chun How Tan for helpful discussions and suggestions.

\appendix

\section{Variance of Information Gain}

When the quality $q$ of a user is uncertain, then the information gain for each
question is also uncertain. Under the assumption the probability $q$, that a
user answers a question correctly, is the same across all questions, and that
the prior is a uniform distribution, then the variance of the information gain
distribution is given by:

\begin{align}
Var[IG(q,n)] =  \log(n)^2 + \frac{2ab  \cdot I_{ab} }{(a+b)(a+b+1)} \nonumber  \\
 + \frac{a(a+1) \cdot J_a }{(a+b+1)(a+b)}  + \frac{b(b+1)  \cdot J_b}{(a+b+1)(a+b)}    \nonumber\\
  + \frac{b(b+1) \log(n-1)^2 }{(a+b)(a+b+1)} +  \frac{2b  \cdot \log(n-1) \cdot K_{ab}}{(a+b) \cdot (a+b+1)}  \nonumber\\
  -  \frac{2\log(n) \cdot \log(n-1) \cdot b }{a+b} - 2\log(n) \cdot E[IG(a,b,n)]   \nonumber
\end{align}
\begin{eqnarray}
J_a & = & (\Psi(a+2) - \Psi(a+b+2))^2 \nonumber\\
	& &	 + \Psi_1(a+2) - \Psi_1(a+b+2)\nonumber\\
J_b & = & (\Psi(b+2) - \Psi(a+b+2))^2 \nonumber\\
	& &	 + \Psi_1(b+2) - \Psi_1(a+b+2)\nonumber\\
I_{ab} & = & (\Psi(a+1) - \Psi(a+b+2)) \cdot (\Psi(b+1) \nonumber \\
       &   &  - \Psi(a+b+2)) - \Psi_1(a+b+2) \nonumber \\
K_{ab} & = &  (a+b+1) \Psi(a+b+1)  \nonumber \\
       &   & -(b+1)\Psi(b+1) -a \Psi(a+1)\nonumber
\end{eqnarray}

\noindent where $\Psi(x)$ is the digamma function,  $\Psi_1(x)$ is the trigamma
function,  $n$ is the number of options presented to the user, $a-1$ is the
number of correct, and $b-1$ is the number of incorrect answers submitted by the
user~\cite{archer2013entropy}.
 
{\scriptsize
%\bibliographystyle{acm}
%\bibliography{references}

}

\end{document}